# Improved Diagnosis of Tibiofemoral Cartilage Defects on MRI Images Using Deep Learning

Gergo Merkely M.D. [1], Alireza Borjali Ph.D. [2,3], Molly Zgoda [1], Evan M. Farina M.D. [1], Simon Görtz M.D. [1], Orhun Muratoglu Ph.D. [2], Christian Lattermann M.D. [1], Kartik M. Varadarajan Ph.D. [2]

[1] Dept. Orthopaedic Surgery, Division of Sports Medicine, Center for Cartilage Repair, Brigham and Women's Hospital, Harvard Medical School, Boston, MA, USA

[2] Department of Orthopaedic Surgery, Harris Orthopaedics Laboratory, Massachusetts General Hospital, Boston, MA, USA

[3] Department of Orthopaedic Surgery, Harvard Medical School, Boston, MA, USA

**Gergo Merkely M.D. and Alireza Borjali Ph.D. are both first authors.**

**Corresponding author:**

Gergo Merkely M.D., Department Orthopaedic Surgery, Division of Sports Medicine, Center for Cartilage Repair, Brigham and Women's Hospital, Harvard Medical School, 850 Boylston Street 112, Chestnut Hill, MA, 02467, USA.

**Email:** gmerkely@bwh.harvard.edu

**Abstract:**

**Background:** MRI is the modality of choice for cartilage imaging, however, its diagnostic performance is variable and significantly lower than the gold standard diagnostic knee arthroscopy. In recent years, deep learning has been used to automatically interpret medical images to improve diagnostic accuracy and speed.

**Purpose:** The primary purpose of this study was to evaluate whether deep learning applied to the interpretation of knee MRI images can be utilized to accurately identify cartilage defects. The secondary purpose of this study was to compare deep learning's performance in identifying cartilage defects on standard MRI sequences to those of an orthopaedic trainee and those of an experienced orthopaedic surgeon.

**Methods:** We analyzed data from patients who underwent knee MRI evaluation and consequently had arthroscopic knee surgery (207 with cartilage defect, 90 without cartilage defect). Patients' arthroscopic findings were compared to preoperative MRI images to verify the presence or absence of isolated tibiofemoral cartilage defects. For each patient, the most representative MRI image slice of the patient's condition was selected (defect or no-defect) from the coronal view and from the sagittal view based on the arthroscopic findings. We developed three convolutional neural networks (CNNs) to analyze the images: CNN-1 trained on the images of the sagittal and coronal views; CNN-2 trained on the images of the sagittal view; CNN-3 trained on the images of the coronal view. We implemented image-specific saliency maps to visualize the CNNs decision-making process. To compare the CNNs' performance against human interpretation, the same test dataset images were provided to an experienced orthopaedic surgeon and an orthopaedic trainee.

**Results:** Saliency maps demonstrated that the CNNs learned to focus on the clinically relevant areas of the tibiofemoral articular cartilage on MRI images during the decision-making processes. The CNN-1 achieved higher performance than the orthopaedic surgeon, with two more accurate diagnoses made by the CNN-1. The orthopaedic surgeon outperformed the CNN-2 and CNN-3. The orthopaedic trainee had the worst overall performance.

**Conclusion:** CNN can be used to enhance the diagnostic performance of MRI in identifying isolated tibiofemoral cartilage defects and may replace diagnostic knee arthroscopy in certain cases in the future.

**What is known about this subject:**

Although the utilization of MRI is considered the standard of care in identifying cartilage defects, the accuracy of MRI in detecting these lesions varies and therefore, improvement of its reliability with respect to chondral lesions is needed. Deep learning is a subset of machine learning that is mainly concerned with image analysis and extracting knowledge from complex imaging data sets such as medical images. Radiologists and orthopaedic surgeons have previously applied deep learning strategies to provide automatic interpretations of medical images to improve their diagnostic accuracy and speed.

**What this study adds to existing knowledge:**

In this study, we demonstrate that a trained convolutional neural network can be utilized in clinical settings to improve the speed and diagnostic accuracy of MRI interpretation to identify cartilage defects compared to manual review by an expert. Furthermore, these experiments show that with the provision of additional data sources during training, CNNs can integrate these additional data points to arrive at more accurate outputs.

**Introduction**

Articular cartilage injuries are common and have the potential to progress to osteoarthritis if left untreated.[13] In the clinical work-up of a suspected symptomatic chondral defect, magnetic resonance imaging (MRI) is the modality of choice to better assess such pathology[11, 17]. This non-invasive approach allows articular cartilage to be better observed since its high soft-tissue contrast displays a different signal intensity for articular cartilage compared to the nearby menisci and bone[6, 14, 16]. Although the utilization of MRI is considered the standard of care, the accuracy of MRI in detecting these lesions varies depending on several factors including the MRI technique, protocol, and magnet strength as well as the size of the lesion[23]. Furthermore, articular cartilage can be a challenging tissue to image due to its very thin and layered microarchitecture overlying a complex 3D osseous base[12]. There is a wide reported range of diagnostic performance of 2D FSE MR for assessing knee cartilage, with overall sensitivity ranging from 26%–96%, specificity of 50%–100%, and accuracy of 49%–94%[8, 15, 18, 19]. Previous studies have demonstrated that deeper lesions, such as Grade III and IV, are identified more often while smaller/shallower or earlier stage lesions are not as accurately detected[10]. Detecting early cartilage degeneration is crucial to the treatment and prevention of further symptomatic pain, as well as in reducing risk for and delaying progression of OA[9]. Following MRI, the next diagnostic step often entails diagnostic knee arthroscopy before determination of a definitive intervention[11]. While arthroscopy is the gold standard in diagnosing chondral lesions, its expense and invasive nature drives the quest for more accurate pre-operative screening systems that may one day obviate the role of diagnostic arthroscopy prior to definitive surgery. Currently, standard MRI may allow for this but improvement of its reliability with respect to chondral lesions is needed.

In recent years, machine learning has gained rapid popularity in medical applications, revolutionizing the way high-volume medical data is processed and interpreted.[2] In general, machine learning refers to a series of mathematical algorithms that enable the machine to "learn" the relationship between input and output data without being explicitly told how to do so[2]. Deep learning is a subset of machine learning that is mainly concerned with image analysis and extracting knowledge from complex imaging data sets such as medical images[1, 2, 4]. Radiologists and orthopaedic surgeons have previously applied deep learning strategies to provide automatic interpretations of medical images to improve their diagnostic accuracy and speed[20, 22, 19]. In addition, deep learning may be used to transfer the expertise of tertiary centers' to smaller community institutions and more remote areas where access to such experts is scarce.

Consequently, this study's primary aim was to evaluate whether deep learning applied to the interpretation of MRI images and founded upon a "ground truth" of arthroscopically-verified diagnoses can be utilized to identify cartilage defects accurately. The secondary aim was to compare deep learning's performance in identifying cartilage defects to those of an experienced orthopaedic surgeon and a less experienced orthopaedic resident.

**Methods**

*Data collection*

After institutional review board approval, informed consent was obtained from all patients when they were entered into our study database. In this retrospective study, we analyzed data from patients who underwent knee MRI evaluation and consequently had arthroscopic knee surgery between September 2011 and December 2019. Proton density-weighted fast spin-echo MRI scans (1.5-T; Siemens) were used. Patients' preoperative MRI images were compared to arthroscopic findings by two orthopaedic surgeons to verify whether an isolated cartilage defect was present in the tibiofemoral joint or not. In particular, as a first step arthroscopy images and surgical notes were assessed by two independent orthopedic surgeons and the presence or absence of isolated tibiofemoral cartilage defects was identified. After the defect status was confirmed arthroscopically, one image from the coronal view and one image from the sagittal view that were the most representative imaging slices of the patient's condition (defect or no-defect) were selected from the entire preoperative MRI series for each patients by the two orthopedic surgeons independently. Disagreement on the most representative image was resolved by discussion among the two examiner. The senior investigator was consulted in situations where disagreement persisted.

Two-hundred and ninety-seven knees were evaluated in this study, 207 with cartilage defect and 90 without cartilage defect in the tibiofemoral joint. Patients with cartilage defects were significantly older (37.6 ± 11.2) than patients with no defect (33.8 ± 11.8) ($p<0.05$). In addition, more females had cartilage defects (120 females had a defect [58% of all defects] vs. 36 females with no cartilage defect [40% of all no-defects], $p<0.05$). Baseline demographic information is displayed in Table 1. The mean time between MRI and surgery was 3.1 ± 3.4 months.

**Table 1. Patient Demographics**

|  | Total | Defect | No-defect | (95% CI) | P-value |
|---|---|---|---|---|---|
| **Number of patients (%)** | 297 | 207 | 90 | | |
| **Age, y, mean SD** | 36.4 ± 11.5 | 37.6 ± 11.2 | 33.8 ± 11.8 | (0.9 – 6.6) | 0.01 |
| **BMI (kg/m$^2$), mean SD** | 28.1 ± 5.4 | 28.2 ± 5.5 | 28.1 ± 5.2 | (-1.3 - 1.4) | 0.93 |
| **Female, n (%)** | 156 (52.5) | 120 (58.0) | 36 (40.0) | | <0.01 |
| **Right knee, n (%)** | 142 (47.8) | 98 (47.3) | 44 (48.9) | | 0.89 |

SD, standard deviation; y, year; n, number; CI, confidence interval.

*Data analysis*

Images were randomly assigned to "training", "validation", and final "test" subsets with an 80:10:10 split ratio. SPSS (version 21.0; IBM Corp) was used to analyze our patient population's baseline demographics. Continuous variables are reported as mean ± standard deviation, whereas categorical variables are reported as numbers and percentages. The normal distribution of the data was confirmed using the Shapiro-Wilk test. Continuous data were compared with the independent sample t-test. Categorical data were compared with the Chi-square test. We used the training subset to train three convolutional neural networks (CNNs). The first network (CNN-1) was simultaneously trained on the images of the sagittal and coronal views. The second network (CNN-2) was only trained on the images of the sagittal view, and the third network (CNN-3) was only trained on the images of the coronal view (Figure 1).

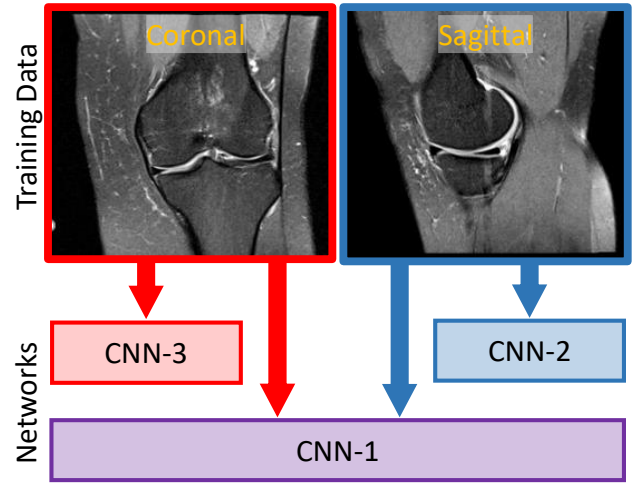

**Figure 1** Three implemented convolutional neural networks (CNNs): 1) CNN-1 simultaneously trained on the images of the sagittal and coronal views, 2) CNN-2 trained only on the images of the sagittal view, and 3) CNN-3 trained only on the images of the coronal view

We optimized hyper-parameters iteratively on the validation dataset using a grid search strategy. We utilized the transfer learning method by modifying a CNN that was initially developed for non-medical image classification and used it for this application[3]. We used Xception CNN architecture[7] pre-trained on ImageNet[5] database for each view to train CNN-2 and CNN-3. Then we used a support vector machine (SVM) to combine two views to train CNN-1. We implemented image-specific saliency maps to visualize the CNNs decision-making process. We used the test dataset (29 patients), which had been kept separate from all previous training and validation processes, to evaluate the CNNs' ultimate performance. These CNNs were implemented using Tensorflow (Keras backend) on a workstation comprised of an Intel(R) Xeon(R) Gold 6128 processor, 64GB of DDR4 RAM, and an NVIDIA Quadro P5000 graphic card.

To compare the CNNs' performance against human interpretation, the same test dataset images were also provided to two independent clinicians including an experienced orthopaedic surgeon and a less experienced orthopaedic resident. It is worth mentioning that the orthopaedic surgeon had extensive expertise in the diagnosis and treatment of cartilage defects. The orthopaedic surgeon and the orthopaedic resident had access to both the sagittal and coronal view images, while they were blinded to the arthroscopic findings. In all test analyses, the orthopaedic surgeon and the orthopaedic resident, as well as all the CNNs were asked to identify whether a cartilage defect was present or absent in a binary fashion.

**Results**

Saliency maps demonstrated that CNNs applied focused consideration along the clinically relevant articular cartilage margins of the tibiofemoral joint on MRI images during the decision-making processes. (Figure 2)

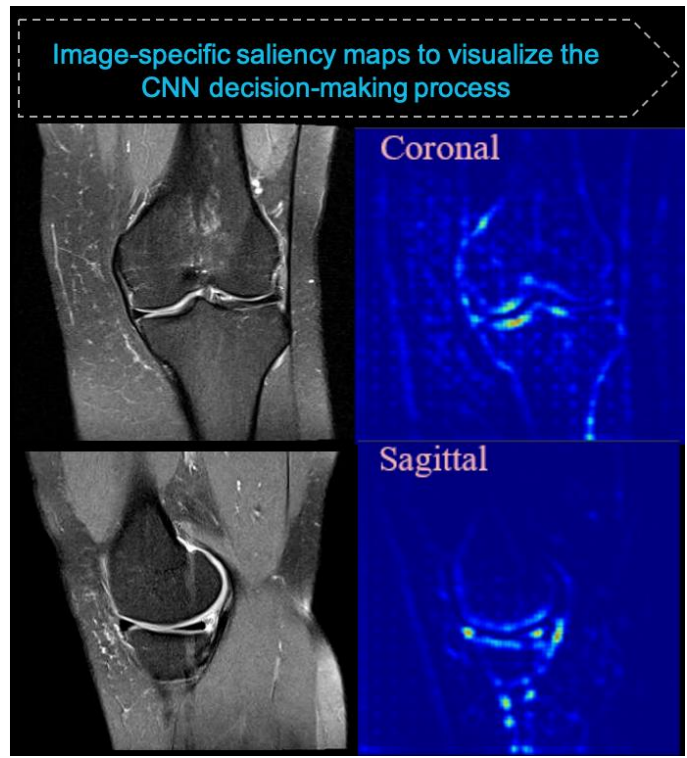

**Figure 2** The saliency maps of a representative patient's MRI, coronal, and sagittal views. Colored regions in the saliency maps, where red denotes higher relative influence than blue, indicate the most influential regions on the CNN's performance.

Figure 3 shows the receiver operating characteristic (ROC) curves for all CNNs classifying the MRI images into "defect" and "no defect" categories. The diagnostic performances of the orthopaedic surgeon and the orthopaedic resident are also overlaid in Figure 3. While all CNNs achieved the same area under the curve value, at the reported threshold, CNN-1 outperformed CNN-2 and CNN-3.

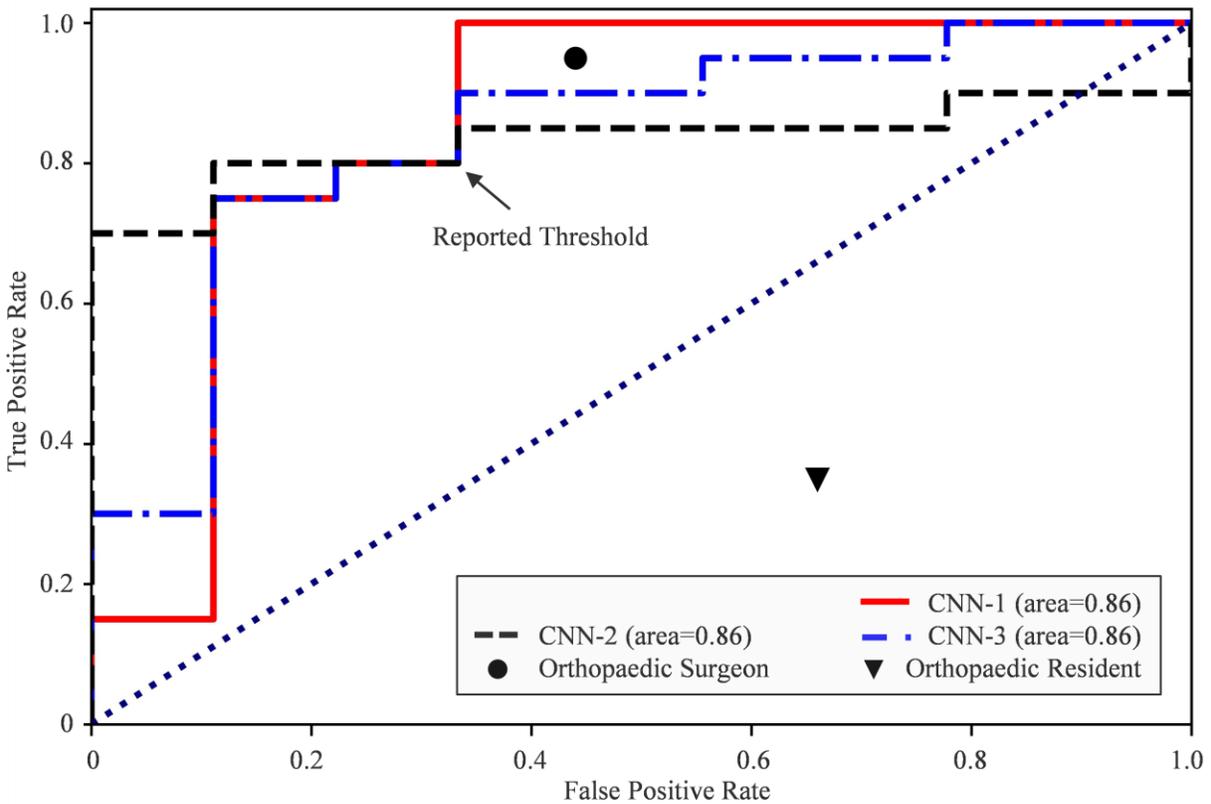

**Figure 3** Receiver operating characteristic (ROC) curve for the CNNs with the orthopaedic surgeon and orthopaedic resident results superimposed.

Table 2 shows the binary diagnoses (cartilage defect vs. no-defect) of all the CNNs and the orthopaedic surgeon and the orthopaedic resident for the 29 patients in the test dataset.

**Table 2** Diagnosis results of the orthopaedic surgeon, the orthopaedic resident, and all the CNNs for patients in the test dataset

| Patient # | Ground Truth | Orthopaedic Surgeon | Orthopaedic Resident | CNN-1 | CNN-2 | CNN-3 |
|---|---|---|---|---|---|---|
| 1 | defect | defect | defect | defect | defect | no-defect |
| 2 | defect | defect | no-defect | defect | no-defect | defect |
| 3 | defect | defect | no-defect | defect | no-defect | defect |
| 4 | defect | defect | no-defect | defect | defect | defect |
| 5 | defect | defect | no-defect | defect | defect | defect |
| 6 | defect | defect | defect | defect | defect | defect |
| 7 | defect | defect | defect | defect | defect | defect |
| 8 | defect | defect | no-defect | defect | defect | defect |
| 9 | defect | defect | no-defect | defect | defect | defect |
| 10 | defect | defect | defect | defect | defect | no-defect |
| 11 | defect | defect | defect | defect | no-defect | defect |
| 12 | defect | defect | defect | defect | defect | defect |
| 13 | defect | no-defect | no-defect | defect | defect | defect |
| 14 | defect | defect | no-defect | defect | defect | defect |
| 15 | defect | defect | no-defect | defect | defect | defect |
| 16 | defect | defect | no-defect | defect | defect | defect |
| 17 | defect | defect | no-defect | defect | defect | defect |
| 18 | defect | defect | no-defect | defect | defect | defect |
| 19 | defect | defect | defect | defect | defect | defect |
| 20 | defect | defect | no-defect | defect | defect | defect |
| 21 | no-defect | defect | defect | no-defect | no-defect | no-defect |
| 22 | no-defect | no-defect | defect | no-defect | defect | no-defect |
| 23 | no-defect | defect | defect | defect | defect | defect |
| 24 | no-defect | no-defect | defect | no-defect | no-defect | no-defect |
| 25 | no-defect | no-defect | no-defect | defect | no-defect | defect |
| 26 | no-defect | defect | defect | no-defect | defect | no-defect |
| 27 | no-defect | no-defect | no-defect | defect | no-defect | defect |
| 28 | no-defect | defect | no-defect | no-defect | no-defect | no-defect |
| 29 | no-defect | no-defect | defect | no-defect | no-defect | no-defect |

CNN-1 slightly outperformed the orthopaedic surgeon and made two more accurate diagnoses overall, with one more accurate diagnosis of defect and one more accurate diagnosis of no-defect made by the CNN-1. All CNNs significantly outperformed the orthopaedic resident. Interestingly, both the orthopaedic surgeon (sensitivity 82.61%, specificity 83.33%, positive predictive value [PPV] 95%, negative predictive value [NPV] 55.56%) and the CNN-1 (sensitivity 86.96, specificity 100%, PPV 100%, NPV 66.67%), were less accurate in determining a negative diagnosis of no-defect compared to determining a positive diagnosis of present cartilage defect (Table 3).

**Table 3**: Diagnostic performance of the orthopaedic surgeon, the orthopaedic resident, and all the CNNs for patients in the test dataset

|  | Orthopaedic Surgeon | Orthopaedic Resident | CNN-1 | CNN-2 | CNN-3 |
|---|---|---|---|---|---|
| **Accuracy** | 82.76% | 34.48% | 89.66% | 79.31% | 82.76% |
| **Sensitivity** | 82.61% | 53.85% | 86.96% | 85.00% | 85.71% |
| **Specificity** | 83.33% | 18.75% | 100.00% | 66.67% | 75.00% |
| **PPV*** | 95.00% | 35.00% | 100.00% | 85.00% | 90.00% |
| **NPV**** | 55.56% | 33.33% | 66.67% | 66.67% | 66.67% |

* positive predictive value (PPV), **negative predictive value (NPV)

**Discussion**

In this study, we developed three CNNs to provide a binary automated diagnosis for the presence or absence of isolated tibiofemoral cartilage defects utilizing MRI images. We also visualized the decision-making process of these CNNs using saliency maps to highlight the influential regions of MRI images on the CNNs outcome. These saliency maps demonstrated that these CNNs appropriately focused on clinically relevant articular cartilage margins to make a diagnosis (Figure 2). Importantly, these CNNs were not provided any direct instructions on where in the entire MRI image to search for the defect. We showed that the diagnostic accuracy of these CNNs was comparable to that of an experienced orthopaedic surgeon (Table 2). The CNN-1 achieved the best overall performance with an accuracy of 89.66%, sensitivity of 89.96%, specificity of 100%, PPV of 100%, and NPV of 66.67%; ultimately arriving at 2 (out of 29) additional accurate diagnoses over the orthopaedic surgeon (Figure 3). These findings demonstrate that a trained CNN can be utilized in clinical settings to improve the speed and diagnostic accuracy of MRI interpretation to identify cartilage defects compared to manual review by an expert. Furthermore, these experiments show that with the provision of additional data sources during training, CNNs can integrate these additional data points to arrive at more accurate outputs. This is demonstrated by the CNN-1, which was simultaneously trained on the images of the sagittal and coronal views outperforming the CNN-2 and CNN-3, which were trained on the images of sagittal or coronal views alone, respectively.

The reported diagnostic performance of 2D FSE MR for assessing knee cartilage is wide, with overall sensitivity ranging from 26%–96%, specificity of 50%–100%, and accuracy of 49%–94%[8, 15, 18, 19]. A meta-analysis of 8 studies conducted by Zhang et al. concluded that manually reviewed knee MRI demonstrates an overall sensitivity of 75% (95% CI, 62% to 84%), specificity of 94% (95% CI, 89% to 97%), and diagnostic odds ratio of 12.5 (95% CI, 6.5 to 24.2) in detecting knee chondral lesions, higher than grade $I^2$. Similarly, in a cross-sectional study of 36 patients comparing MRI-grading (manual) and arthroscopic-grading of cartilage disease; von Engelhardt et al. calculated grade-specific diagnostic values with sensitivity ranging from 20% to 70%, specificity ranging from 74% to 95%, and overall accuracy ranging from 70% to 92%[21]. Based on the low sensitivities appreciated in both of these studies, the authors conclude that a negative result on MRI should not preclude diagnostic arthroscopy[21, 23]. By comparison, the results of this study suggest a significant improvement in the sensitivity and specificity of automatic CNN-enhanced MRI interpretation over traditional manual review in detecting cartilage defects.

This study's significant strength is its foundation upon a "ground truth" of diagnoses verified by arthroscopy. In the field of machine and deep learning, the "ground truth" is the objective reality that the model gets trained on to predict an outcome or automate a data analysis task. The best objective measure, or "gold standard," for cartilage pathology is direct visualization and interrogation by arthroscopy. As such, our CNNs are trained to directly predict the "gold standard" arthroscopic diagnosis, free of any intervening subjective bias inherent to human radiologic interpretation. Yet we caution that the results of this early study bear limited generalizability at this time. All the CNNs underwent vigorous deep learning protocols incorporating a sizable dataset of 297 subjects, but were "trained" to determine only a binary diagnosis of isolated cartilage "defect" or "no-defect." Conclusions of diagnostic performance surrounding parameters of disease severity, the grade of cartilage damage, or concomitant pathology cannot be made. Similarly, comparisons against human performance are limited to the scope of a single orthopaedic surgeon and a single orthopaedic resident. However, these limitations mark exciting frontiers for further investigation. Furthermore, our study demonstrated that additional input data can be effectively integrated by CNNs to provide more accurate output interpretations. As such, our future work seeks to incorporate full data volumes from 3-view MRI series, a diversity of expert imaging interpretations, more granular diagnostic grades, and cases with patellofemoral or concomitant intra-articular pathology to further hone the diagnostic performance and clinical utility of CNN-enhanced MRI interpretation.

The results from this study are encouraging as the accuracy of these CNNs proves comparable to, if not outperforming that of orthopaedic surgeons. The diagnostic utility of such deep learning networks may indeed be a helpful tool, in the future for clinical application of imaging data. With high sensitivity values approaching 100% demonstrate that CNN-enhanced MRI interpretation may serve as an effective screening tool for the presence or absence of articular cartilage defects. If further scrutinized and validated, this may indeed obviate the need for a diagnostic arthroscopy in many cases.

In addition, deep learning tools like these can be integrated into software and imaging systems, bringing tertiary centers' diagnostic expertise to community hospitals and remote rural clinics.

**Conclusion**

Convolutional neural networks (CNNs) can be used to enhance the diagnostic performance of MRI in identifying isolated tibiofemoral cartilage defects and may guide indications for diagnostic knee arthroscopy in the future.

**Acknowledgment**

No external funding was received for this study.

**References:**

1. Borjali A, Chen A, Bedair H, et al. Comparing performance of deep convolutional neural network with orthopaedic surgeons on identification of total hip prosthesis design from plain radiographs. *medRxiv.* 2020.
2. Borjali A, Chen A, Muratoglu O, Morid M, Varadarajan K. Deep Learning in Orthopedics: How Do We Build Trust in the Machine? *Healthcare Transformation.* 2020;0.
3. Borjali A, Chen AF, Muratoglu OK, Morid MA, Varadarajan KM. Detecting total hip replacement prosthesis design on plain radiographs using deep convolutional neural network. *J Orthop Res.* 2020;38(7):1465-1471.
4. Borjali A, Langhorn J, Monson K, Raeymaekers B. Using a patterned microtexture to reduce polyethylene wear in metal-on-polyethylene prosthetic bearing couples. *Wear.* 2017;392:77-83.
5. Deng J, Dong W, Socher R, Li L-J, Li K, Fei-Fei L. Imagenet: A large-scale hierarchical image database. *In 2009 IEEE conference on computer vision and pattern recognition*; 2009.
6. Disler DG, Peters TL, Muscoreil SJ, et al. Fat-suppressed spoiled GRASS imaging of knee hyaline cartilage: technique optimization and comparison with conventional MR imaging. *AJR Am J Roentgenol.* 1994;163(4):887-892.
7. F. C. Xception: Deep learning with depthwise separable convolutions. *Proceedings of the IEEE conference on computer vision and pattern recognition*; 2017.
8. Figueroa D, Calvo R, Vaisman A, Carrasco MA, Moraga C, Delgado I. Knee chondral lesions: incidence and correlation between arthroscopic and magnetic resonance findings. *Arthroscopy.* 2007;23(3):312-315.
9. Kijowski R, Blankenbaker DG, Munoz Del Rio A, Baer GS, Graf BK. Evaluation of the articular cartilage of the knee joint: value of adding a T2 mapping sequence to a routine MR imaging protocol. *Radiology.* 2013;267(2):503-513.

10. McGibbon CA, Trahan CA. Measurement accuracy of focal cartilage defects from MRI and correlation of MRI graded lesions with histology: a preliminary study. *Osteoarthritis Cartilage.* 2003;11(7):483-493.
11. Merkely G, Ackermann J, Lattermann C. Articular Cartilage Defects: Incidence, Diagnosis, and Natural History. *Oper Tech Sports Med.* 2018;26:156-161.
12. Merkely G, Hinckel B, Shah N, Small K, Lattermann C. Magnetic Resonance Imaging of the Patellofemoral Articular Cartilage. *Patellofemoral Pain, Insatbility and Arthritis*. Berlin, Heidelberg: Springer; 2020:47-61.
13. Messner K, Maletius W. The long-term prognosis for severe damage to weight-bearing cartilage in the knee: a 14-year clinical and radiographic follow-up in 28 young athletes. *Acta Orthop Scand.* 1996;67(2):165-168.
14. Potter HG, Linklater JM, Allen AA, Hannafin JA, Haas SB. Magnetic resonance imaging of articular cartilage in the knee. An evaluation with use of fast-spin-echo imaging. *J Bone Joint Surg Am.* 1998;80(9):1276-1284.
15. Quatman CE, Hettrich CM, Schmitt LC, Spindler KP. The clinical utility and diagnostic performance of magnetic resonance imaging for identification of early and advanced knee osteoarthritis: a systematic review. *Am J Sports Med.* 2011;39(7):1557-1568.
16. Recht MP, Kramer J, Marcelis S, et al. Abnormalities of articular cartilage in the knee: analysis of available MR techniques. *Radiology.* 1993;187(2):473-478.
17. Rodrigues MB, Camanho GL. Mri Evaluation of Knee Cartilage. *Rev Bras Ortop.* 2010;45(4):340-346.
18. Smith TO, Drew BT, Toms AP, Donell ST, Hing CB. Accuracy of magnetic resonance imaging, magnetic resonance arthrography and computed tomography for the detection of chondral lesions of the knee. *Knee Surg Sports Traumatol Arthrosc.* 2012;20(12):2367-2379.
19. Sonin AH, Pensy RA, Mulligan ME, Hatem S. Grading articular cartilage of the knee using fast spin-echo proton density-weighted MR imaging without fat suppression. *AJR Am J Roentgenol.* 2002;179(5):1159-1166.
20. Tiulpin A, Thevenot J, Rahtu E, Lehenkari P, Saarakkala S. Automatic Knee Osteoarthritis Diagnosis from Plain Radiographs: A Deep Learning-Based Approach. *Sci Rep.* 2018;8(1):1727.
21. von Engelhardt LV, Lahner M, Klussmann A, et al. Arthroscopy vs. MRI for a detailed assessment of cartilage disease in osteoarthritis: diagnostic value of MRI in clinical practice. *BMC Musculoskelet Disord.* 2010;11:75.
22. Xue Y, Zhang R, Deng Y, Chen K, Jiang T. A preliminary examination of the diagnostic value of deep learning in hip osteoarthritis. *PLoS One.* 2017;12(6):e0178992.
23. Zhang M, Min Z, Rana N, Liu H. Accuracy of magnetic resonance imaging in grading knee chondral defects. *Arthroscopy.* 2013;29(2):349-356.